\documentclass[10pt, twocolumn, switch]{article} 

\usepackage{preprint}

\usepackage{amsmath, amsthm, amssymb, amsfonts}
\usepackage{indentfirst}

\usepackage[numbers,square]{natbib}
\usepackage[utf8]{inputenc}	
\usepackage[T1]{fontenc}	
\usepackage{xcolor}		
\usepackage[colorlinks = true,
            linkcolor = purple,
            urlcolor  = blue,
            citecolor = cyan,
            anchorcolor = black]{hyperref}	
\usepackage{booktabs} 		
\usepackage{nicefrac}		
\usepackage{microtype}		
\usepackage{lineno}		
\usepackage{float}			

\usepackage{lipsum}		

\usepackage{newfloat}
\DeclareFloatingEnvironment[name={Supplementary Figure}]{suppfigure}
\usepackage{sidecap}
\sidecaptionvpos{figure}{c}

\usepackage{titlesec}
\titlespacing\section{0pt}{12pt plus 3pt minus 3pt}{1pt plus 1pt minus 1pt}
\titlespacing\subsection{0pt}{10pt plus 3pt minus 3pt}{1pt plus 1pt minus 1pt}
\titlespacing\subsubsection{0pt}{8pt plus 3pt minus 3pt}{1pt plus 1pt minus 1pt}

\usepackage{tikz,xcolor,hyperref}

\definecolor{lime}{HTML}{A6CE39}
\DeclareRobustCommand{\orcidicon}{
	\begin{tikzpicture}
	\draw[lime, fill=lime] (0,0) 
	circle [radius=0.16] 
	node[white] {{\fontfamily{qag}\selectfont \tiny ID}};
	\draw[white, fill=white] (-0.0625,0.095) 
	circle [radius=0.007];
	\end{tikzpicture}
	\hspace{-2mm}
}
\foreach \x in {A, ..., Z}{\expandafter\xdef\csname orcid\x\endcsname{\noexpand\href{https://orcid.org/\csname orcidauthor\x\endcsname}
			{\noexpand\orcidicon}}
}

\usepackage{amsmath}

\title{Partial Visual-Semantic Embedding: Fashion Intelligence System\\with Sensitive Part-by-Part Learning}

\usepackage{authblk}

\author[1,2\thanks{\tt{shi3mizu8-r@fuji.waseda.jp}}]{Ryotaro Shimizu}
\author[2]{Takuma Nakamura}
\author[1]{Masayuki Goto}

\affil[1]{Waseda University}
\affil[2]{ZOZO Research}

\begin{document}

\twocolumn[ 
  \begin{@twocolumnfalse} 
  
\maketitle

\begin{abstract}
In this study, we propose a technology called the Fashion Intelligence System based on the visual-semantic embedding (VSE) model to quantify abstract and complex expressions unique to fashion, such as ``casual,'' ``adult-casual,'' and ``office-casual,'' and to support users' understanding of fashion. However, the existing VSE model does not support the situations in which the image is composed of multiple parts such as hair, tops, pants, skirts, and shoes. We propose partial VSE, which enables sensitive learning for each part of the fashion coordinates. The proposed model partially learns embedded representations. This helps retain the various existing practical functionalities and enables image-retrieval tasks in which changes are made only to the specified parts and image reordering tasks that focus on the specified parts. This was not possible with conventional models. Based on both the qualitative and quantitative evaluation experiments, we show that the proposed model is superior to conventional models without increasing the computational complexity.
\end{abstract}
\vspace{1.0cm}

  \end{@twocolumnfalse} 
] 


\section{Introduction}
\label{sec:intro}
When browsing fashion items online, users must interpret fashion images to resolve difficult questions that arise in their minds, without the shopkeeper's support. The questions are as follows: 1) ``what would this outfit look like if it were more formal?''; 2) ``how office-casual is this outfit?''; and 3) ``what makes this outfit street?''. It is difficult to answer these questions even for experts and particularly difficult for non-experts. This ambiguity inherent in the fashion field may hinder the users' from pursuing their deep interest in the fashion industry, making it difficult for them to try new genres of clothing. Therefore, we expect that automatically obtaining answers to these questions will arouse interest among users and broaden their perceptions, thereby helping them interpret fashion clothing.

In response to this expectation, Shimizu et al.~\cite{Shimizu2022_FashionIntelligenceSystem} proposed a ``Fashion Intelligence System'' to support the interpretation of these terms through various applications that can be provided by applying a visual-semantic embedding (VSE) model. This system helps users obtain answers to ambiguous questions by clarifying the relationships between the full-body outfit images and various expressions, including abstract expressions specific to the fashion domain. Specifically, answers to abstract questions from users (in particular, the abovementioned questions 1--3) can be obtained by embedding a full-body outfit image and tag information attached to the image in the same projective space and utilizing the embedded representation of the image and tags in this projective space. Additionally, by enabling users to obtain the answers, it is possible to reduce the ambiguity inherent in fashion and support the user in all fashion-related decisions and actions such as what to wear and what items to purchase.

\begin{figure}[ht]
\centering
\includegraphics[width=1.0\linewidth]{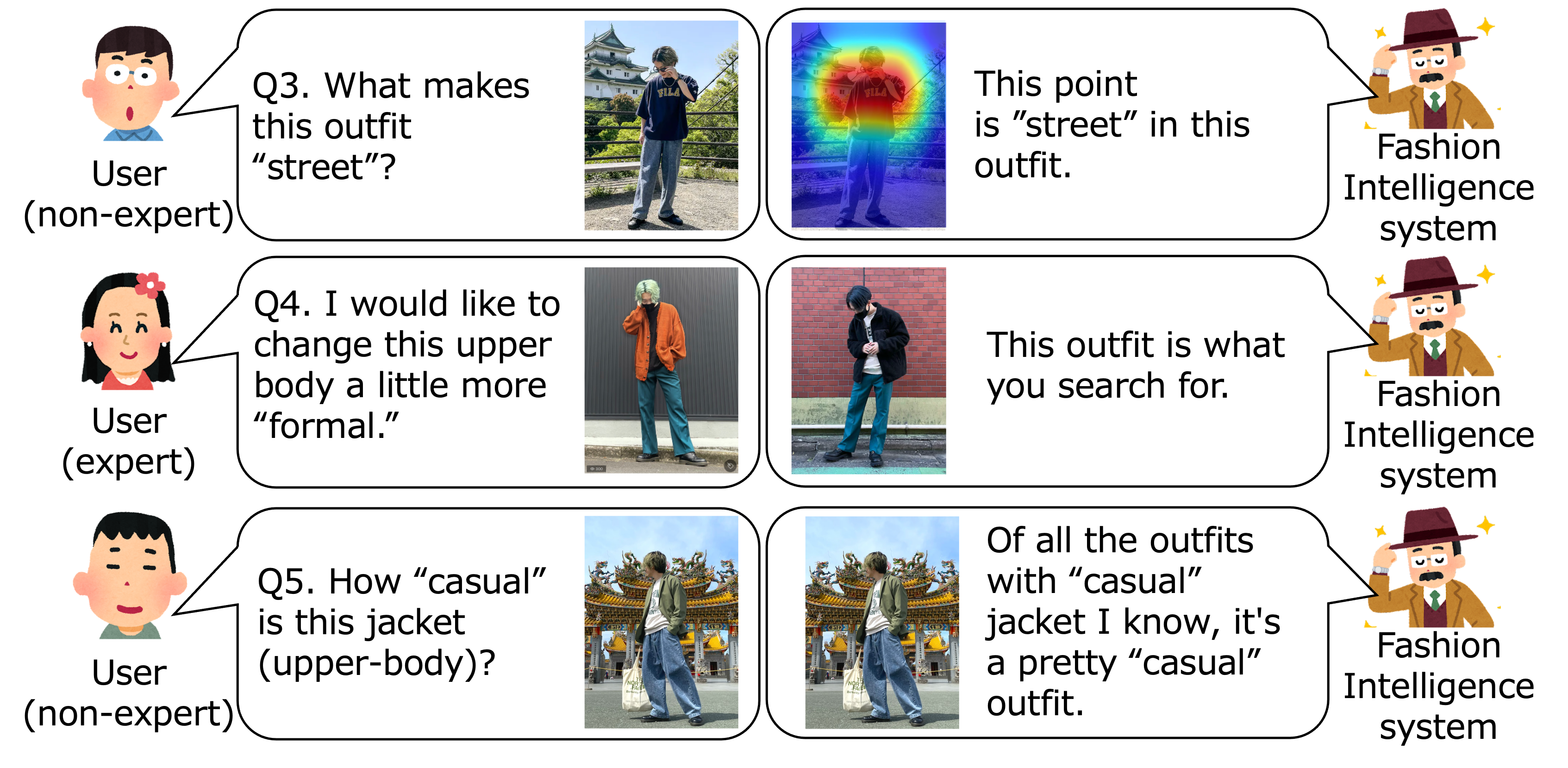}
\caption{Image of fashion intelligence system \label{fig_fis_image}}
\end{figure}

A full-body outfit image consists of many elements such as hair, tops, pants, skirts, and shoes. Specifically, a full-body outfit image can be considered to be a set of these components. However, the VSE in~\cite{Shimizu2022_FashionIntelligenceSystem} has a simple model structure that maps the full-body outfit image to the projective space as a batch. Furthermore, obtaining the embedded representation corresponding to each part is one of its biggest limitations. The problem caused by this limitation is that the model cannot answer the questions which users truly want to be answered: 4) ``what would the outfit look like if I changed the upper body to make it a little more formal?''; and 5) ``how casual is this jacket?'' Specifically, image-retrieval results should not show images in which changes are made to the entire body~\cite{Wei-Lin_fashionplusplus} because users consider how to dress based on the outfits they already have.

In this study, we propose a partial VSE (PVSE) model that enables the acquisition of an embedded representation corresponding to each part in a full-body outfit image while maintaining a simple model structure and a low computational complexity. The proposed model retains various previous practical application functions and enables image-retrieval tasks in which changes are made only to specified parts (answering question 4) and image reordering tasks, attentively focusing on the specified parts (answering question 5). This was not possible with conventional models.
Furthermore, based on multiple quantitative evaluation experiments and qualitative evaluation analyses using real-world service data, we show that the proposed model has superior functionality compared to conventional models.

The main contributions of this study are as follows. 1) We propose a PVSE model that can map a full-body outfit image and rich tags into the same projective space and obtain an embedded representation corresponding to each part containing the full-body outfit. 2) Evaluation experiments using real-world data show that the simple structure of the proposed model contributes to more accurate mapping operations while maintaining the computational complexity of the conventional VSE model. 3) We show that the proposed model not only maintains various practical application functions to support the user's fashion interpretation (which the conventional model has) but also identifies image retrieval and image reordering tasks that attentively focus on a specific part (which cannot be identified by the conventional VSE model) through multifaceted analysis experiments and discussions using real-world data. Consequently, the proposed model reduces the ambiguity and complexity inherent in fashion through various visualization methods and supports users' marketing activities and fashion decisions.

\section{Related Research}
\label{sec:related_research}

\subsection{Visual-Semantic Embedding}
VSE models have been widely researched for image caption and image description retrieval~\cite{VSE_image2text_text2image, VSEpp, VSE_image2text_text2image_2}, visual question-answering~\cite{VSE_QA, VSE_QA2}, image-to-description and description-to-image generation~\cite{VSE_disc_gen, VSE_text-to-image_gen, Tewel_vse_imagecaptioning_2022_CVPR}, hashing tasks~\cite{VSE_hashing, Yang_vse_hashing_2022_CVPR}, person re-identification~\cite{VSE_PersonIdentification}, and video similarity evaluation~\cite{Zeng_vse_videosimirality_2022_CVPR}. In the fashion domain, VSE has been used for text-based and individual clothing image retrieval~\cite{VSE_ClothRetrieval, VSE_ClothRetrieval2} and learning outfit compatibility (individual outfit item matching)~\cite{VSE_ClothMatching, Han_OutfitComposition_withText_2017}. Furthermore, the VSE included in~\cite{HAN2017_FASHIONCONCEPTDISCOVERY} is a method for mapping fashion item images (not full-body outfit images) and specific words (not abstract words) contained in the item descriptions in the same projective space.

A VSE model that allows the mapping of full-body outfit images and abstract rich attributes was proposed as a starting point for research on fashion intelligence systems that support the interpretation of fashion-specific expressions~\cite{Shimizu2022_FashionIntelligenceSystem}.
Additionally, the dual Gaussian VSE (DGVSE) model has been proposed \cite{Shimizu2022_DGVSE}, which assumes a multidimensional Gaussian distribution behind the embedded representations of images and attributes.
However, these models do not include a mechanism for learning a full-body outfit separately. Specifically, it can only capture the entire atmosphere of a full-body outfit.

\subsection{Part-by-Part Learning of Fashion Style Images}
There are studies that independently learn fashion images of each item included in a full-body outfit and recommend items based on the compatibility between different types of items~\cite{song_compatibility_2018, Kang_compatibility_2019, Vasileva_compatibility_2018, Zou_compatibility_2022_CVPR}, studies that derive which combination of candidate (inside a wardrobe) items match~\cite{WeiLin_outfitmatching_2018, Dong_outfitmatching_2019}, studies that search for other items that match a query item~\cite{set2set_matching, Chen_outfitgeneration_2019, Feng_OutfitComposion_2019, Li_OutfitComposition_2017}, and studies that predict future fashion trends for each item~\cite{Ziad_ItemTrendPrediction_2017}.
While these tasks are concerned with which combinations of items are fashionable, the interests of our study include: ``what happens if this casual outfit becomes more casual?'' and ``how casual is this outfit?'' Thus, the focus of this study is fundamentally different. Furthermore, the fact that each item is independently applied to a convolutional neural network (CNN) is also a major difference in our study, which focuses on learning full-body outfit images directly.

Additionally, in SNS data, it is almost impossible for a full-body outfit image (one post) to be linked to the data on individual items, where real trends in fashion have accumulated most recently. Therefore, the method foThe problem caused by this limitation is that the model cannot answer the users' questions which users truly want to be answeredr acquiring the features of each part from a single full-body image, as envisioned in this study, is particularly desirable in situations such as when analyzing SNS data (which are useful for e-commerce site data).
Studies in the field of person re-identification have used segmentation-based methods to extract the features for each part from a single full-body image~\cite{Li2020_reindetification_parts, Guo2019_reindetification_parts}. However, these studies are clearly different from our study in that they include an operation to mix the features of each part because there is no need to correspond the order of the parts to each dimension of the final required features. Furthermore, another method extracts a bounding box for each body part from a single image by detection and performs learning for each part~\cite{Zhao2017_prediction_patch, trip_advisor_mcnn}. However, this detection-based (patch-based) method is less sensitive than the segmentation-based method, and the computational complexity necessary for applying each part to a large network remains a challenge.

Conversely, several approaches to fashion image generation have been studied based on shape features obtained by semantic segmentation and capturing features for each part~\cite{Wei-Lin_fashionplusplus, ViTON, Dong_2020_CVPR, Jiang_ViTON_2022_CVPR}. Particularly, Hsiao et al.~\cite{Wei-Lin_fashionplusplus} performed excellent work based on the claimed that ``making minor changes to fashion is important to become fashionable.'' However, these approaches are based on image generation (i.e., items that do not strictly exist are generated), and the research does not focus on attributes (expressions), but rather on the question, ``what minor changes should be made to become fashionable from the target full-body outfit?''

\section{Methodology}
\subsection{Model Architecture}
\begin{figure}[ht]
\centering
\includegraphics[width=1.\linewidth]{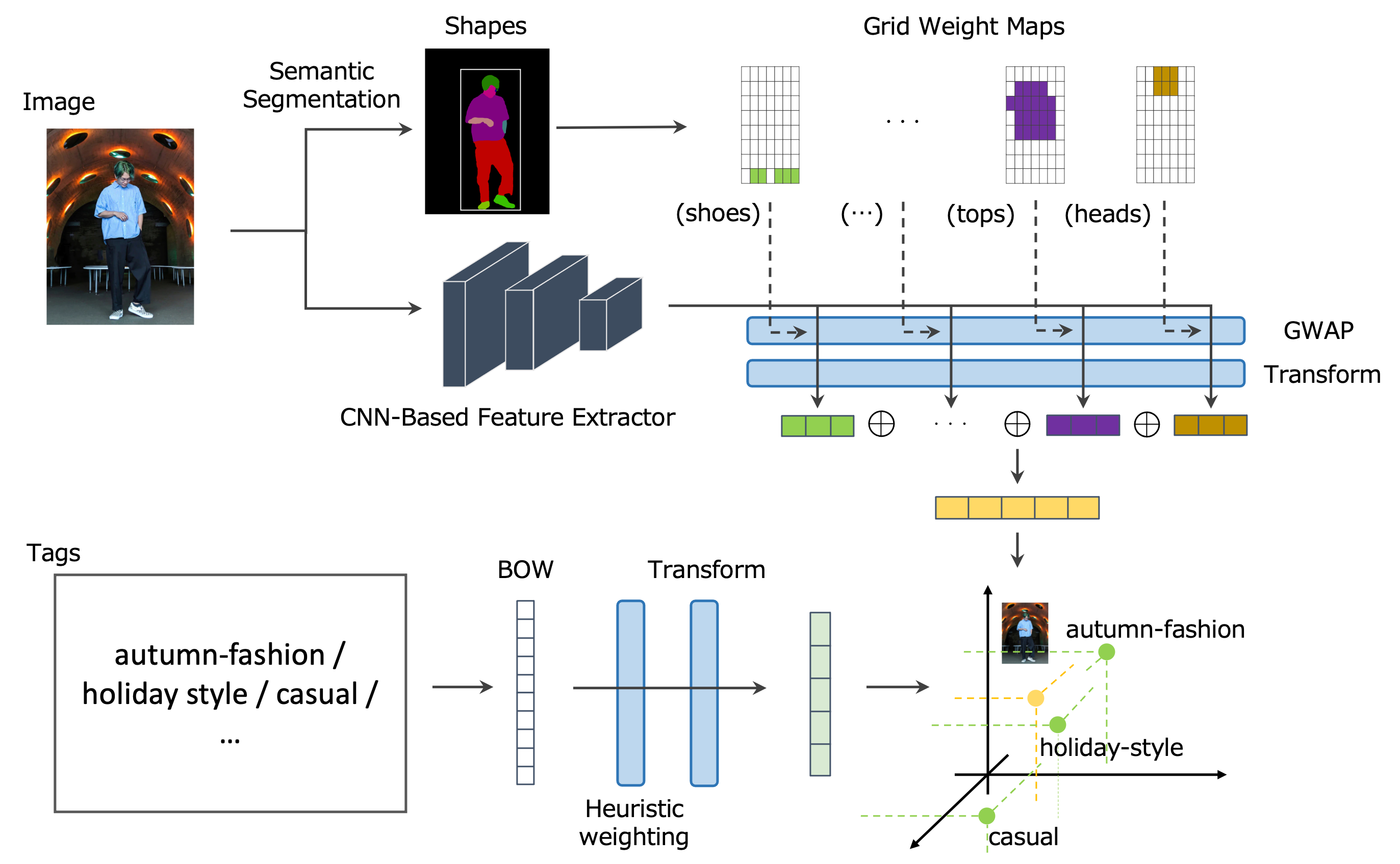}
\caption{Structure of a prototype of our partial visual-semantic embedding model proposal \label{fig_structure}}
\end{figure}

The structure of the entire model is shown in Figure~\ref{fig_structure}.

The key feature of the proposed model is the inclusion of an architecture that considers a full-body outfit image as a collection of fashion items (parts) and acquires the embedded representation corresponding to each part. A simple but effective architecture is obtained by extending foreground-centered learning~\cite{Shimizu2022_FashionIntelligenceSystem} and combining the grid weight map corresponding to each part obtained from the semantic segmentation model and the CNN-based feature with global weighted average pooling (GWAP)~\cite{gwap}. With this method, regardless of the number of parts that the full-body outfit image is divided into, the number of times that the CNN is applied to each image per epoch can be limited to only one time, thus avoiding an increase in computational complexity.

\subsection{Part-by-Part Grid Weight Map Acquisition}
The semantic segmentation included in this study calculates the probability of the fashion item appearing in each pixel. The part with the highest probability for each pixel is considered a part of that pixel. Here, the grid refers to the area in which the target image is divided vertically into $I$ and horizontally into $J$. The grid weight map for the $l$-th part is defined as $G_{l} = \{g_{(1,1), l}, ..., g_{(i,j), l}, ..., g_{(I,J), l}\}$ when the number of all parts is defined as $L$. $N_{(i, j), l}$ is defined as the count of the $l$-th part of the pixels contained in the $(i, j)$-th grid and $g_{(i, j), l} = N_{(i, j), l} / \sum_{i=1}^{I} \sum_{j=1}^{J} N_{(i, j), l}$. In other words, $g_{(i, j), l}$ is the percentage of pixels in the $(i,j)$-th grid out of all the pixels in the $l$-th part. The grid weight map $G_{l}$ was used to achieve delicate learning for each part.

\subsection{Parameter Optimization}
The dataset used in this study consisted of a single full-body outfit image to which multiple tags were assigned. First, the image is embedded using the image features obtained from the CNN and a grid weight map.
Eq.~(\ref{deqn:image_embedding}) is used to obtain a concatenated embedded representation of the features for each part of each image.
\begin{eqnarray}
\label{deqn:image_embedding}
{\rm \mathbf{x}} &=& \left[ {\rm \mathbf{x}}_1 ; \cdots ; {\rm \mathbf{x}}_l ; \cdots ; {\rm \mathbf{x}}_L \right], \\
{\rm \mathbf{x}}_l &=& \sum_{i=1}^{I} \sum_{j=1}^{J} {\rm \mathbf{W}}_{{\rm I},l} {\rm \mathbf{f}}_{(i,j)},
\end{eqnarray}
where $\left[{\rm \mathbf{a}}; {\rm \mathbf{b}} \right]$ is the concatenate operation between vectors ${\rm \mathbf{a}}$ and ${\rm \mathbf{b}}$, ${\rm \mathbf{x}} \in \mathbb{R}^{KL}$ is the embedded representation (vertical vector) of the full-body outfit image, and ${\rm \mathbf{x}}_l \in \mathbb{R}^{K}$ is the embedded representation (vertical vector) of the $l$-th fashion item part in the image. $K$ is the number of dimensions of the embedded representation for each part. ${\rm W}_{\rm I} = \{ {\rm \mathbf{W}}_{{\rm I},1},\cdots,{\rm \mathbf{W}}_{{\rm I},l},\cdots,{\rm \mathbf{W}}_{{\rm I},L} | {\rm \mathbf{W}}_{{\rm I},l} \in \mathbb{R}^{D \times K} \}$ is a set of transformation matrices for mapping image features (vertical vector) of the $(i,j)$-th grid ${\rm \mathbf{f}}_{(i,j)} \in \mathbb{R}^{D}$ obtained from a CNN into the projection space, where $D$ is the number of dimensions of the final convolutional layer of the CNN. This operation makes it possible to proceed with subsequent learning based on an understanding of the parts that correspond to each dimension of the embedded representation to be acquired. All the vectors defined in this study are vertical vectors unless specified otherwise.

Furthermore, the embedded representation of the tag set assigned to an image is heuristically weighted to generate an embedded representation considering the bias in the frequency with which each tag is assigned to the entire dataset. The heuristic weighting rule is based on the assertion that ``tags appearing infrequently in the overall dataset are more likely to be important elements that characterize the image (differentiate it from other images).''
\begin{eqnarray}
\label{deqn_tag_embedding}
{\rm \mathbf{v}} &=& \sum_{t=1}^{T} w_{t} {\rm \mathbf{v}}_t, \\
w_{t} &=& \frac{1 / \mathrm{log}(N_{t}+1)}{\sum_{t=1}^{T} 1 / \mathrm{log}(N_{t}+1)},
\end{eqnarray}
where ${\rm \mathbf{v}} \in \mathbb{R}^{KL}$ is the embedded representation of the tag set, ${\rm \mathbf{v}}_t \in \mathbb{R}^{KL}$ is the embedded representation of the $t$th single tag, $N_t$ indicates the total attachment frequency of the $t$th attached tag to the target image in the entire mini-batch, and $T$ is the total number of tags included in the target image.

By optimizing Eq.~(\ref{deqn_loss}), which includes the abovementioned features, the full-body outfit image and the attached tags are mapped into the same projective space, and the embedded representation for each part corresponding to the full-body outfit image and the embedded representation for the tags are obtained.
\begin{align}
\label{deqn_loss}
&l_{\rm npair\&ang}(\mathrm{O}) = \notag \\
&\hspace{1mm} l_{\rm npair}(\mathrm{O}) + \lambda \Biggl( \frac{1}{2N} \sum^{N}_{n=1} \log \Bigl(1+ \sum_{m \neq n} \exp \{f_{\rm ang}({\rm \mathbf{x}}, {\rm \mathbf{v}}, {\rm \mathbf{v}}_m)\} \Bigr) \notag \\
&\hspace{3mm} + \frac{1}{2N} \sum^{N}_{n=1} \log \Bigl(1+\sum_{m \neq n} \exp \{f_{\rm ang}({\rm \mathbf{v}}, {\rm \mathbf{x}}, {\rm \mathbf{x}}_m)\} \Bigr) \biggr),
\end{align}
\vspace{-3mm}
\begin{align}
&l_{\rm npair}(\mathrm{O}) = \notag \\
&\hspace{3mm} \frac{1}{2N} \sum^{N}_{n=1} \log \Bigl(1+\sum_{m \neq n} \exp \{{\rm \mathbf{x}}^\top {\rm \mathbf{v}}_m - {\rm \mathbf{x}}^\top {\rm \mathbf{v}}\} \Bigr) \notag \\
&\hspace{3mm} + \frac{1}{2N} \sum^{N}_{n=1} \log \Bigl(1+\sum_{m \neq n} \exp \{{\rm \mathbf{v}}^\top {\rm \mathbf{x}}_m - {\rm \mathbf{v}}^\top {\rm \mathbf{x}}\} \Bigr),
\end{align}
where $\mathrm{O} = \{ {\rm V}, {\rm W}_{\rm I}, {\rm \mathbf{W}}_{\rm T} \}$ is a set of target parameters to be optimized, ${\rm V}$ is a parameter set contained in the CNN, ${\rm \mathbf{W}}_{\rm T} \in \mathbb{R}^{H \times KL}$ is the transform matrix from a bag-of-words representation to the $t$-th tag-embedded representation ${\rm \mathbf{v}}_t$, and $H$ is the number of unique tags in the entire dataset.
Additionally, $N$ is the number of positive samples in the batch data, and $\{ {\rm \mathbf{e}}_{\rm anc}, {\rm \mathbf{e}}_{\rm pos}, {\rm \mathbf{e}}_{\rm neg} \}$ are the anchor, positive, and negative samples (embedded representations), respectively. Furthermore, $\lambda$ is a positive hyperparameter that compensates for the n-pair loss~\cite{npair_loss} and angular loss~\cite{angular_loss}, and $\alpha$ is the angular loss margin (angle). Furthermore, each embedded representation is normalized when calculating the loss.

\begin{figure}[ht]
\centering
\includegraphics[width=1.\linewidth]{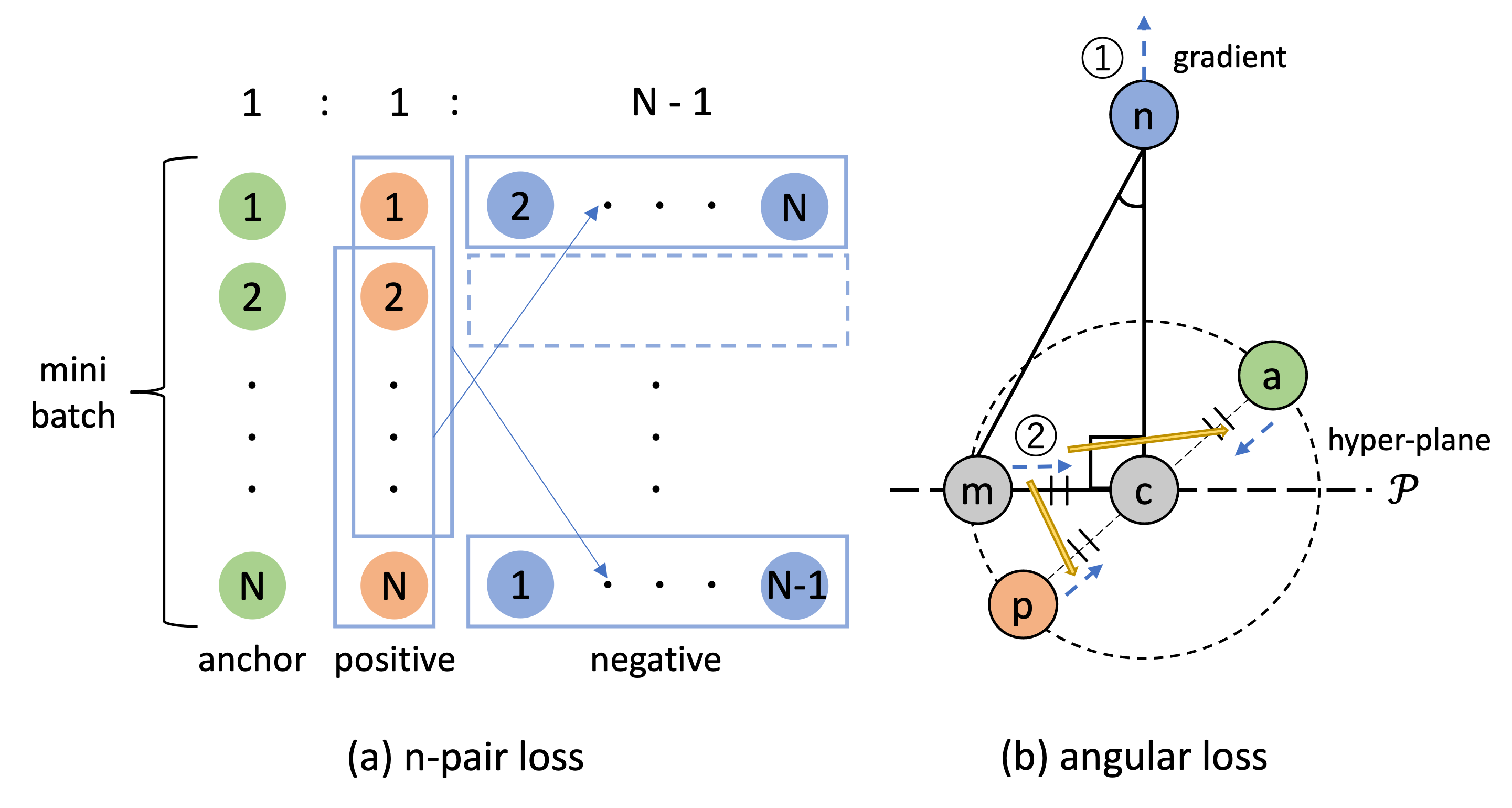}
\caption{Images of n-pair loss \& angular loss \label{fig_img_loss}}
\end{figure}

The loss function is defined by a combination of n-pair loss and angular loss, which is more stable than triplet loss~\cite{triplet_loss} employed in many VSE models. The loss was calculated, as shown in Figure~\ref{fig_img_loss}.

In the n-pair loss, all the positive samples in the mini-batch, except for the positive sample corresponding to the target anchor sample, are treated as negative samples and trained to move away from the anchor sample. This system allows us to utilize numerous samples in a single training session without increasing the computational complexity, thereby achieving stable learning.

Furthermore, angular loss considers the relative positional relationship (angle) between the anchor and positive and negative samples to achieve stable learning. Specifically, as shown in Figure~\ref{fig_img_loss}(b), triangle $\triangle {\rm cmn}$ is structured by 1) midpoint c (coordinate vector ${\rm \mathbf {e}}_{\rm c}$) between anchor point a (coordinate vector ${\rm \mathbf{e}}_{\rm anc}$) and positive point p (coordinate vector ${\rm \mathbf{e}}_{\rm pos}$); 2) negative point n (coordinate vector ${\rm \mathbf{e}}_{\rm neg}$); and 3) point m (coordinate vector ${\rm \mathbf{e}}_{\rm m} $) on hyperplane $\mathcal{P}$ perpendicular to edge nc and on the circumference of the circle of radius ac(cp) centered at point c
and is utilized to achieve learning by considering the relative positions of the anchor, positive, and negative. The basic idea is as follows: by making the angle $\angle {\rm cnm}$ smaller than the margin $\alpha$, the gradient works in two directions (1 and 2 in Figure~\ref{fig_img_loss}). The negative sample moves away from the anchor sample, and the positive sample moves closer. This concept is expressed through trigonometric functions, as expressed by Eq.~(\ref{eq:tangent}).
\begin{eqnarray}
\label{eq:tangent}
\tan \angle {\rm cnm} = \frac{||{\rm \mathbf{e}}_{\rm m} - {\rm \mathbf{e}}_{\rm c}||}{||{\rm \mathbf{e}}_{\rm neg} - {\rm \mathbf{e}}_{\rm c}||} = \frac{||{\rm \mathbf{e}}_{\rm anc} - {\rm \mathbf{e}}_{\rm pos}||}{2||{\rm \mathbf{e}}_{\rm neg} - {\rm \mathbf{e}}_{\rm c}||} \leq \tan \alpha,
\end{eqnarray}
where $||{\rm \mathbf{e}}_{\rm m} - {\rm \mathbf{e}}_{\rm c}|| = \frac{||{\rm \mathbf{e}}_{\rm anc} - {\rm \mathbf{e}}_{\rm pos}||}{2}$ is established, because the edge cm is half of the diameter ap.
Furthermore, Eq.~(\ref{eq:tangent}) is expanded in Eq.~(\ref{eq:tangent_plus}).
\begin{align}
\label{eq:tangent_plus}
&f_{\rm ang}({\rm \mathbf{e}}_{\rm anc},{\rm \mathbf{e}}_{\rm pos},{\rm \mathbf{e}}_{\rm neg}) \nonumber \\
&\hspace{1mm}= ||{\rm \mathbf{e}}_{\rm anc} - {\rm \mathbf{e}}_{\rm pos}||^2 - 4||{\rm \mathbf{e}}_{\rm neg} - {\rm \mathbf{e}}_{\rm c}||^2 \tan^2 \alpha \nonumber \\
&\hspace{1mm}= 4({\rm \mathbf{e}}_{\rm anc} + {\rm \mathbf{e}}_{\rm pos})^\top {\rm \mathbf{e}}_{\rm neg} \tan^2 \alpha - 2 {{\rm \mathbf{e}}_{\rm anc}}^\top {\rm \mathbf{e}}_{\rm pos} (1+\tan^2 \alpha),
\end{align}
where the coordinates of point c are expressed as ${\rm \mathbf{e}}_{\rm c} = \frac{||{\rm \mathbf{e}}_{\rm anc} + {\rm \mathbf{e}}_{\rm pos}||}{2}$, and the constant terms that depend on the value of ${\rm \mathbf{e}}$ are dropped in the process of unfolding. Eq.~(\ref{deqn_loss}) was derived by extending this angular loss to n pairs (batch angular loss) and combining it with the n-pair loss.

\section{Experimental Evaluation}
To quantitatively confirm the effectiveness of the proposed PVSE model, two types of evaluation experiment were conducted using the accumulated data in the fashion coordination posting application WEAR~\cite{app_wear}.

\subsection{Experimental Settings}
\label{sec_experiment_setting}
The total number of full-body outfit images in the experimental data was 15,740, and the number of unique tags attached to all the images was 1104. Additionally, all the participants reflected in the target images were female.
The dimensions of the embedded representation included in the PVSE model were set as $KL$ = 128. A stochastic gradient descent (SGD) optimizer was used with an initial learning rate of 0.01 and was halved every five epochs. The overall number of epochs was 50, batch size was 32, and margin $\alpha$ was set to 36°. Furthermore, we used GoogleNet Inception v3~\cite{inception_v3} pre-trained on ImageNet~\cite{deng2009imagenet}. Moreover, the context embedding with edge perceiving (CE2P) framework~\cite{CE2P} (backbone of the future extractor: ResNet-101~\cite{resnet101}) with self-correction for human parsing (SCHP) strategy~\cite{SCHP} pre-trained on LIP~\cite{LIP_dataset} was used for semantic segmentation~\cite{semantic_segmentation_implementation}.

\subsection{Each Component's Position in Projection Space}
\label{sec:evaluation_1}
The validity of the obtained embedding representation is verified by whether the image and tag that should be nearby are mapped together closely. For example, an image with a ``casual'' tag is quantitatively evaluated based on the idea that it should be mapped near a ``casual'' tag. Specifically, an image dataset was created by combining images that actually had the target tag and 10 times as many images that did not have the tag. Subsequently, the top-$M$=\{5, 15\} images from the embedded representations of the images in the created image dataset that are mapped particularly closely to the embedded representation of the target tag are retrieved. It was determined whether the target tag had actually been assigned to each of these retrieved images. Precision and NDCG were used as the accuracy measures (P@$M$ and N@$M$, respectively). As a comparison model, we tested the eight methods listed in Table~\ref{table:result_topk}. Here, (f) represents ``foreground-centered learning''~\cite{Shimizu2022_FashionIntelligenceSystem}.
In addition, because text data are not the focus of this study, positional embedding is removed from transformer-VSE (TVSE)~\cite{transformer_vse},
and $H$ in TVSE-$H$ represents the number of heads in the MHA layer. Furthermore, to check the change in the accuracy of the proposed model depending on the division of the parts, we tested the following three models: 1) PVSE-4: proposal model that divided the full-body outfit image into four parts \{Head, Upper-body, Lower-body, Shoes\} and trained; 2) PVSE-8: proposal model with the setting of eight parts \{Head, Upper-body, Dress, Coat, Lower-body, Arm, Leg, Shoes\}; and 3) PVSE-16: proposal model with the setting of eight parts \{Hat, Hair, Glove, Sunglasses, Upper-body, Dress, Coat, Socks, Pants, Jumpsuits, Skirt, Face, Arm, Leg, Left-shoe, Right-shoe\}.
Furthermore, the experiment was repeated 30 times, and a t-test was performed between the results of the proposed models and the model with the highest accuracy among the comparison models. The significance level was set to 5\%.

\begin{table}[ht]
\centering
\caption{Summary of model type evaluation values for top-$M$ images selected using similarity for each tag (average values of 30 times)}
\label{table:result_topk}
\scalebox{0.72}{
\begin{tabular}{c|rrr|rrr}
\hline
 &
  \multicolumn{1}{c}{P@5} &
  \multicolumn{1}{c}{P@10} &
  \multicolumn{1}{c}{P@15} &
  \multicolumn{1}{|c}{N@5} &
  \multicolumn{1}{c}{N@10} &
  \multicolumn{1}{c}{N@15} \\ \hline \hline
Random & 0.091 & 0.085 & 0.086 & 0.085 & 0.085 & 0.086 \\ \hline
VSE~\cite{HAN2017_FASHIONCONCEPTDISCOVERY} & 0.462 & 0.435 & 0.418 & 0.424 & 0.426 & 0.421 \\
VSE(f)~\cite{Shimizu2022_FashionIntelligenceSystem} & 0.483 & 0.451 & 0.428 & 0.441 & 0.443 & 0.432 \\
GVSE~\cite{gvse} & 0.276 & 0.276 & 0.278 & 0.250 & 0.262 & 0.268 \\
DGVSE~\cite{Shimizu2022_DGVSE} & 0.244 & 0.244 & 0.244 & 0.220 & 0.231 & 0.235 \\
TVSE-4~\cite{transformer_vse} & 0.195 & 0.199 & 0.198 & 0.176 & 0.188 & 0.191 \\
TVSE-8~\cite{transformer_vse} & 0.212 & 0.209 & 0.210 & 0.194 & 0.201 & 0.204 \\
TVSE-16~\cite{transformer_vse} & 0.197 & 0.198 & 0.201 & 0.176 & 0.186 & 0.191 \\ \hline
PVSE-4 & \textbf{0.833}$^{**}$ & \textbf{0.771}$^{**}$ & \textbf{0.714}$^{**}$ & \textbf{0.760}$^{**}$ & \textbf{0.759}$^{**}$& \textbf{0.733}$^{**}$ \\
PVSE-8 & 0.799$^{**}$ & 0.738$^{**}$ & 0.694$^{**}$ & 0.728$^{**}$ & 0.726$^{**}$ & 0.707$^{**}$ \\
PVSE-16 & 0.801$^{**}$ & 0.738$^{**}$ & 0.690$^{**}$ & 0.732$^{**}$ & 0.729$^{**}$ & 0.706$^{**}$ \\ \hline
\end{tabular}
}
\end{table}

From Table~\ref{table:result_topk}, the proposed models are more accurate than the comparison models that include conventional VSE models, regardless of how the parts are divided.
Furthermore, the high accuracy of PVSE suggests that the heuristic weighting works well, at least for the target problem.
Additionally, the accuracy varies depending on how the parts are divided. In this experimental case, the rule that divides the parts into \{Head, Upper-body, Lower-body, Shoes\} shows the best accuracy. It can be concluded that the proposed model map is more effective and sensitive compared to the conventional method because all the dimensions of the conventional methods have semantic representations of all parts. This result suggests the effectiveness of the proposed model, which learns by mapping each dimension of the embedded representation to a single part. Additionally, it is intuitively clear that too fine a division of parts does not lead to better accuracy. Specifically, excessively fine partitioning yields an embedded representation of an image for a situation in which many parts are missing (e.g., dress and coat are not worn at the same time). Whether this is a desirable situation is questionable. In situations where the number of dimensions $KL$ is fixed, the larger the number of parts, the smaller is the number of dimensions that represent the information of the important parts included. Therefore, it is important to set an appropriate number of parts and divide the information obtained from the full-body outfit image to the extent that important information is not lost in the mapping process.

\subsection{Attention to Appropriate Parts}
\label{sec:evaluation_2}
It is possible to determine which regions in an image and tag are highly relevant by calculating the relevance score for each grid and tag using the VSE model. Here, we check whether the relevance score is high for an appropriate region. Specifically, when the relevance scores are calculated between tags such as ``t-shirt,'' ``jeans,'' and ``sneakers,'' and the images to which these tags are attached, the relevance scores should be higher for the regions containing ``Upper-body,'' ``Lower-body,'' and ``Shoes,'' respectively.
Based on this idea, we compared the relevance scores calculated between a specific tag and each image attached to the tag, and the top $M$=5 regions were obtained. True labels of the regions were based on the results of semantic segmentation, and precision and NDCG were used as accuracy indices.
The comparison models same as those of the target models were applied in section~\ref{sec:evaluation_1}.
Furthermore, as specific tags for the experiments, five tags that clearly corresponded to each of the four categories of \{Head: (beret, glasses, hair bun, bob hair, knit hat), Upper-body: (one-piece dress, blouse, cardigan, t-shirt, outer), Lower-body: (denim, wide-pants, skirt, pants, black skinny), and Shoes (ballet shoes, Converse, sneakers, sandals, loafers)\} were selected in order of their attached frequency in the entire dataset.

\begin{table}[ht]
\centering
\caption{Summary of model type evaluation values for top-$M$ regions selected using similarity for each tag}
\label{table:result_topk_aam}
\scalebox{0.71}{
\begin{tabular}{c|rr|rr|rr|rr}
\hline
\multicolumn{1}{l}{} &
  \multicolumn{2}{|c}{Head} &
  \multicolumn{2}{|c}{Upper-body} &
  \multicolumn{2}{|c}{Lower-body} &
  \multicolumn{2}{|c}{Shoes} \\
 &
  \multicolumn{1}{|c}{P@5} &
  \multicolumn{1}{c}{N@5} &
  \multicolumn{1}{|c}{P@5} &
  \multicolumn{1}{c}{N@5} &
  \multicolumn{1}{|c}{P@5} &
  \multicolumn{1}{c}{N@5} &
  \multicolumn{1}{|c}{P@5} &
  \multicolumn{1}{c}{N@5} \\ \hline \hline
Random   & 0.161          & 0.145          & 0.547          & 0.494          & 0.346          & 0.306          & 0.096          & 0.087          \\ \hline
VSE      & 0.518          & 0.472          & 0.624          & 0.557          & 0.874          & 0.790          & 0.095          & 0.081          \\
VSE(f)   & 0.690          & 0.627          & 0.850          & 0.770          & 0.854          & 0.774          & 0.134          & 0.121          \\
GVSE     & 0.510          & 0.461          & 0.805          & 0.722          & 0.885          & 0.800          & 0.260          & 0.246          \\
DGVSE    & 0.532          & 0.482          & \textbf{0.874}          & \textbf{0.786}          & 0.869          & 0.789          & 0.109          & 0.110          \\
TVSE-4 & 0.030          & 0.022          & 0.340         & 0.294           & 0.160          & 0.137          & 0.078          & 0.071          \\
TVSE-8 & 0.059          & 0.049          & 0.408         & 0.357           & 0.182          & 0.161          & 0.103          & 0.091          \\
TVSE-16 & 0.055          & 0.048          & 0.435         & 0.382           & 0.194          & 0.169          & 0.114          & 0.102          \\\hline
PVSE-4  & 0.668          & 0.618          & 0.746          & 0.665          & 0.988          & 0.893          & 0.546          & 0.532          \\
PVSE-8  & 0.643          & 0.589          & 0.611          & 0.558          & \textbf{0.991} & \textbf{0.895} & \textbf{0.565} & \textbf{0.540} \\
PVSE-16 & \textbf{0.779} & \textbf{0.742} & 0.764          & 0.691          & 0.983          & 0.888          & 0.480          & 0.463 \\ \hline
\end{tabular}
}
\end{table}
The results illustrated in Table~\ref{table:result_topk_aam} show that the proposed model has better accuracy for most of the indices compared to the comparison model, including the conventional VSE models; only for the upper body, the DGVSE indices are higher, while all the other parts have lower accuracy. This suggests that the DGVSE model is the result of learning to associate each dimension with the upper body. In other words, it is not a desirable learning method to handle all the parts precisely. However, the proposed model is universally accurate in all cases. This is the result of learning each part delicately, suggesting that the proposed model can learn in a way that satisfies the objective of this study.

\section{Additional Analysis}
The experimental evaluation shows that the proposed model can map a full-body outfit image and the attached tags into the same projective space more reasonably than the comparison model. This section demonstrates the effectiveness of the PVSE model as a fashion intelligence system with multiple types of practical applications based on the results obtained from the proposed model. The analysis conditions are the same as those in subsection~\ref{sec_experiment_setting}, and this section includes the aspect of qualitative evaluation.

\subsection{Image Retrieval and Reordering}
\begin{figure*}[ht]
\centering
\includegraphics[width=1.0\linewidth]{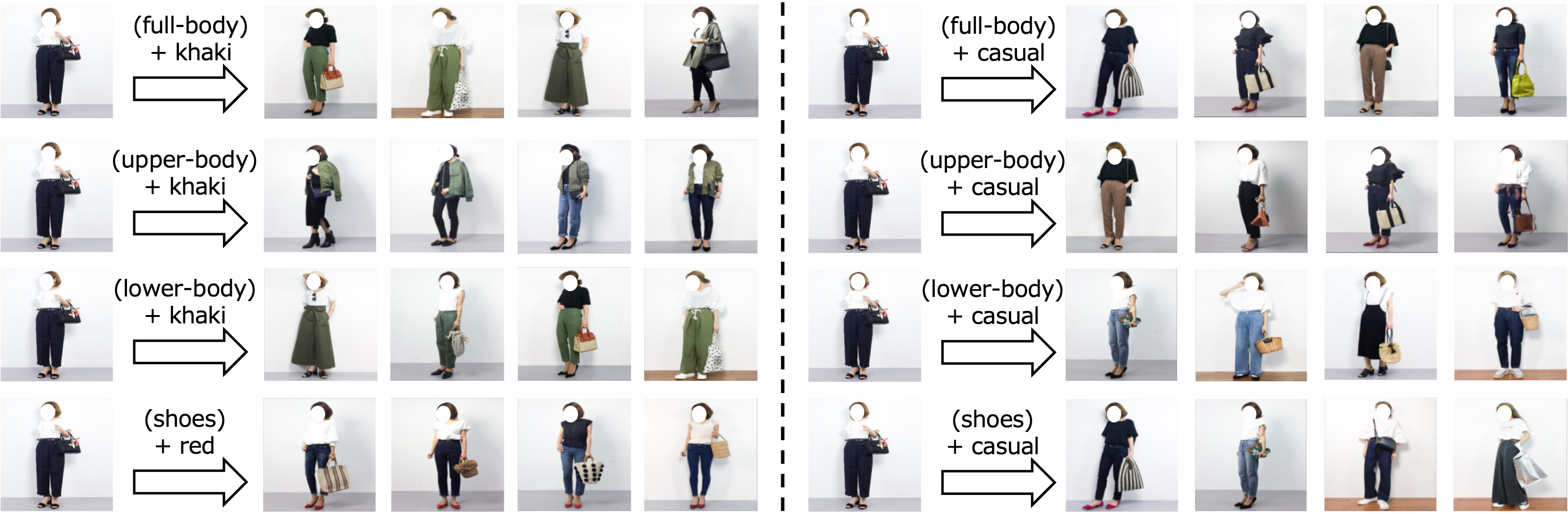}
\caption{Example of image retrieval \label{fig_result_retrieval}}
\end{figure*}

Examples of image retrieval obtained by image and tag operations are shown in Figure~\ref{fig_result_retrieval}.

First, we checked the validity of the results by observing search results with specific tags. When the embedded representation of the ``khaki'' tag is added to the entire embedded representation of the query image (casual adult attire consisting of a white t-shirt and navy pants), an outfit in which the upper or lower body turns khaki in color is returned. In contrast, specifying the parts that have been changed returned outfits in which those parts have been changed (i.e., minor changes have been made). These results indicate that the proposed model can acquire an embedded representation for each part and retrieve the image by making minor changes around the specified part. Furthermore, it is possible to grasp the dressing method that makes a query attire casual using abstract tags. For example, to change the full-body outfit, we can make the upper-body black and add a colorful item to make it more casual. Furthermore, to make a minor change to the lower body and make the overall atmosphere casual, a change from navy skinny to jeans or loose skirts can be made. In this manner, the results obtained from the proposed model can be used to answer ambiguous questions unique to fashion that are difficult for non-experts (and not easy for experts).

\begin{figure*}[ht]
\centering
\includegraphics[width=1.0\linewidth]{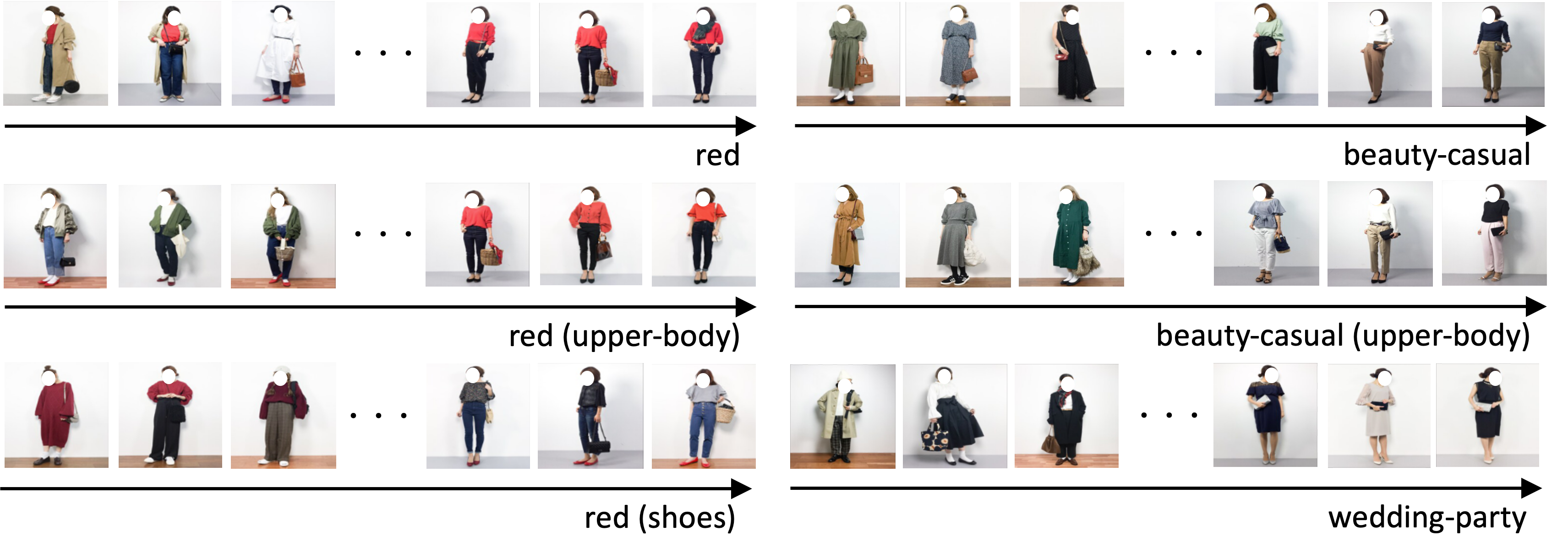}
\caption{Example of image reordering \label{fig_result_reordering}}
\end{figure*}

The results of image reordering obtained by image and tag operations are shown in Figure~\ref{fig_result_reordering}.

First, we checked the validity of the results by observing the sorting results by specific tags. When the sorting with a ``red'' tag was limited to the upper-body and shoes, the images with red items in the specified parts ranked higher. Therefore, the results can be confirmed as reasonable to some extent. Additionally, it is possible to sort by abstract tags. For example, ``beauty-casual'' clothes with a thin silhouette are more ``beauty-casual'' than those with a loose silhouette. Furthermore, it can be understood that typical clothes for a wedding are dresses and those besides dresses are unusual.
If a user wants to go with a unique outfit, it is preferable to choose an outfit with a low relevance score with the ``wedding-party'' tag; if user wants to go with a typical outfit, it's preferable to choose an outfit with a high relevance score.
In this way, it is possible to rearrange images by specifying attention parts to meet the detailed needs of the user, to discover the typical full-body outfits indicated by abstract tags, and to discover suitable clothing for the situation.

\subsection{Attribute Activation Map}
\begin{figure}[ht]
\centering
\includegraphics[width=1.0\linewidth]{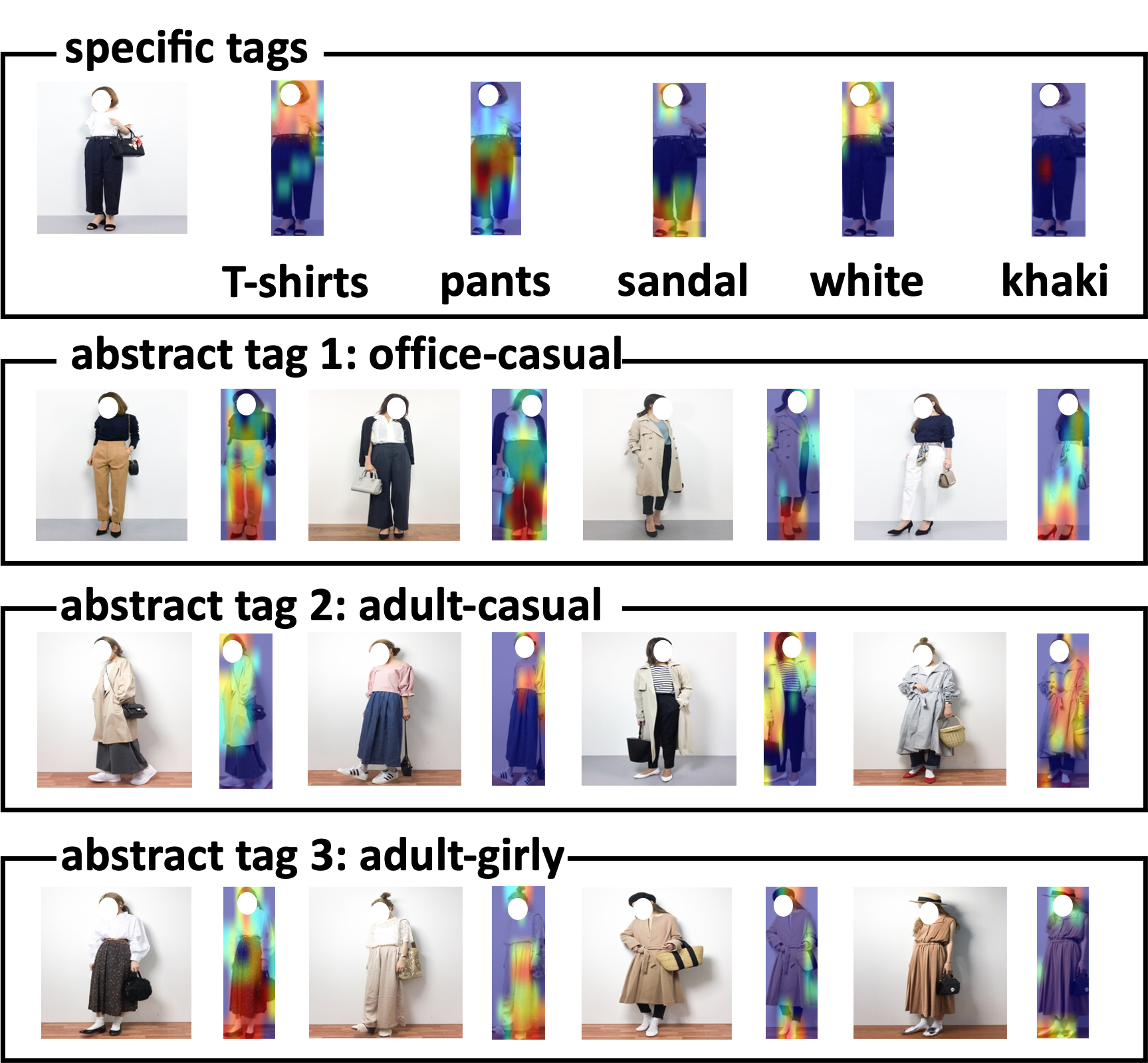}
\caption{Example of AAM \label{fig_result_aam}}
\end{figure}

An attribution activation map (AAM) can be created by calculating the relevance scores of the embedded representation for each region of the image, the representation of the target tag, and representing it in a heatmap. Examples of the AAM are shown in Figure~\ref{fig_result_aam}.

First, we verified the validity of the proposed model by observing the results for specific tags. The results show that ``t-shirts,'' ``pants,'' ``sandals,'' and ``white'' tags, which are actually attached to the target image, are colored in the appropriate places on the image. In contrast, for tags, such as ``khaki,'' which are not relevant to the target fashion image, the relevance score was not high for any of the regions in the image. This indicates that the results are reasonable. Furthermore, when we look at the abstract tags, for example, the upper-body items tend to be key points for ``adult-casual'' coordinates. Additionally, ``adult-girly'' tends to be associated with rounded items; however, in the case of coordinates that include items such as berets and straw hats, these items are the key points. In this way, it is possible to find the region of interest in the full-body outfit image by applying the results of the proposed model.

\section{Discussion}
\subsection{On the Model Structure}
\begin{figure}[ht]
\centering
\includegraphics[width=1.0\linewidth]{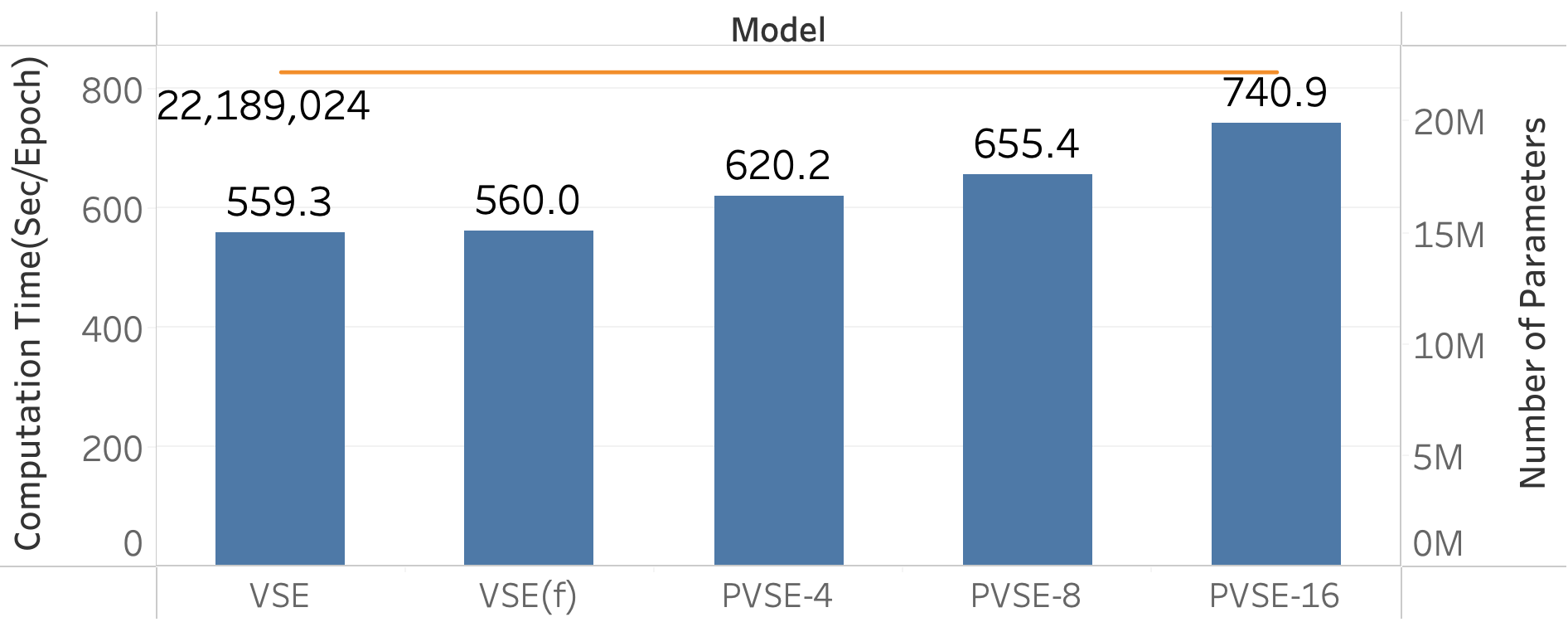}
\caption{Summary of computational amount (time and number of parameters) \label{fig_result_computational_complexity}}
\end{figure}

Figure~\ref{fig_result_computational_complexity} shows the results of comparing the time complexity and space complexity of each model evaluated in the experimental evaluation section.

From the results illustrated in Figure~\ref{fig_result_computational_complexity}, the space computational complexity does not increase compared to the conventional VSE model, regardless of how finely the parts are divided. In other words, the proposed model can be trained regardless of the memory specifications. Furthermore, because the number of parameters did not increase, the number of required training data did not increase. Furthermore, the time complexity increases by approximately 10-30\%. These results are primarily because the CNN does not need to be applied to each item in the full-body image separately.

Under the experimental conditions of this study, the number of CNN parameters accounted for more than 98\% of the entire VSE model, and it accounts for the majority of the total training time for the entire model, even in the case where the forward propagation of the CNN is performed only once for each image. Thus, clearly, a structure in which CNNs are applied independently to each part requires significantly more computation time than an additional cost of 10-30\%. This suggests that our proposed method achieves per-part learning with a minimal increase in computation time. This will lead to a decrease in the throughput and will be of great benefit when considering real-world services.

\subsection{On the Loss Function}
This study adopted the n-pair angular loss when training the proposed model. However, many VSE models use triplet loss~\cite{HAN2017_FASHIONCONCEPTDISCOVERY, Shimizu2022_FashionIntelligenceSystem, gvse, Shimizu2022_DGVSE, VSE_PersonIdentification, VSE_ClothRetrieval, VSEpp, VSE_image2text_text2image} and n-pair loss~\cite{Yang_vse_hashing_2022_CVPR, transformer_vse}, and it is necessary to clarify the accuracy of other types of loss functions to demonstrate the validity of n-pair angular loss. Table~\ref{table:discussion_topk} lists the results of the evaluation experiments for each loss function.

\begin{table}[ht]
\centering
\caption{Summary of loss function type evaluation values for top-$M$ images selected using similarity for each tag (average values of 30 times)}
\label{table:discussion_topk}
\scalebox{0.71}{
\begin{tabular}{c|rrr|rrr}
\hline 
  & \multicolumn{1}{c}{P@5} &
  \multicolumn{1}{c}{P@10} &
  \multicolumn{1}{c}{P@15} & \multicolumn{1}{|c}{N@5} &
  \multicolumn{1}{c}{N@10} &
  \multicolumn{1}{c}{N@15} \\ \hline \hline
triplet & 0.494 & 0.445 & 0.424 & 0.459 & 0.448 & 0.438 \\
n-pair   & 0.593 & 0.558 & 0.526 & 0.544 & 0.548 & 0.534 \\
single angular & 0.130 & 0.125 & 0.124 & 0.119 & 0.121 & 0.123 \\
batch angular & 0.814 & 0.752 & 0.703 & 0.745 & 0.742 & 0.719 \\ \hline 
n-pair angular & \textbf{0.831}$^{**}$ & \textbf{0.768}$^{**}$ & \textbf{0.712}$^{**}$ & \textbf{0.760}$^{**}$ & \textbf{0.758}$^{**}$ &  \textbf{0.731}$^{**}$ \\ \hline
\end{tabular}
}
\end{table}

The results in Table~\ref{table:discussion_topk} show that, compared to the other comparison loss functions, the n-pair angular loss adopted in this study for training the proposed model exhibits the best accuracy. Although omitted for space reasons, the results of the experimental evaluation, as in subsection~\ref{sec:evaluation_2}, also show that the n-pair angular loss was the most effective. N-pair angular loss is a loss function that combines the n-pair loss and batch angular loss by hyperparameter $\lambda$. Comparing the results of single and batch angular loss, the batch angular loss is much higher, suggesting that it is necessary to use a large number of negative samples for a single anchor sample to make the trigonometric idea of angular loss more powerful. Additionally, the batch angular loss was very high when combined with n-pair loss, inspired by \cite{angular_loss}. This suggests that the proposed model can be more robust when combined with learning from both the n-pair and angular perspectives because it requires simultaneous mapping of a complex tag set that includes rich abstract tags and a complex image consisting of a combination of many parts.

\section{Conclusions and Limitations}
In this study, we proposed a PVSE model that can obtain an embedded representation for each of the multiple parts included in the full-body outfit image and attached rich tags. Each dimension of the resulting embedded representation of the image and tags corresponded to a single part of the full-body outfit. This feature of the proposed model maintains the three functions feasible in the previous VSE model and adds two new functions. Additionally, we confirmed that the proposed model outperforms the conventional model in terms of the accuracy of multiple types of evaluation experiments. Furthermore, we confirmed that these advantages can be obtained with little increase in the computational complexity. The proposed model is expected to be used in real-world applications such as supporting the users' purchasing activities on an e-commerce site and users' browsing activities and learning of fashion knowledge on SNSs.

The points that have not been clarified regarding the necessity and effectiveness of this study include the fact that when transforming the tag set to the embedded representation, heuristic weighting is used. The results of the quantitative and qualitative evaluations on this dataset are reasonable; however, it would be ideal if the heuristic part could be eliminated in the model training algorithm.

\section*{Competing Interests}
The authors declare no conflicts of interest.

\section*{Acknowledgements}
This work was supported by JSPS KAKENHI Grant Number 21H04600.

\normalsize

\clearpage

\appendix
\renewcommand{\thesection}{\Alph{section}}
\renewcommand{\thesubsection}{\Alph{section}.\arabic{subsection}}
\setcounter{section}{0} 
\renewcommand{\theequation}{\Alph{section}.\arabic{equation}}
\setcounter{equation}{0}
\renewcommand{\thefigure}{\Alph{section}.\arabic{figure}}
\setcounter{figure}{0}
\renewcommand{\thetable}{\Alph{section}.\arabic{table}}
\setcounter{table}{0}

\section*{Appendix}

\section{Problem Definition}
An example of a full-body clothing image and its tags are shown in Figure~\ref{fig_sample_dataset} below.

\begin{figure}[ht]
\centering
\includegraphics[width=1.0\linewidth]{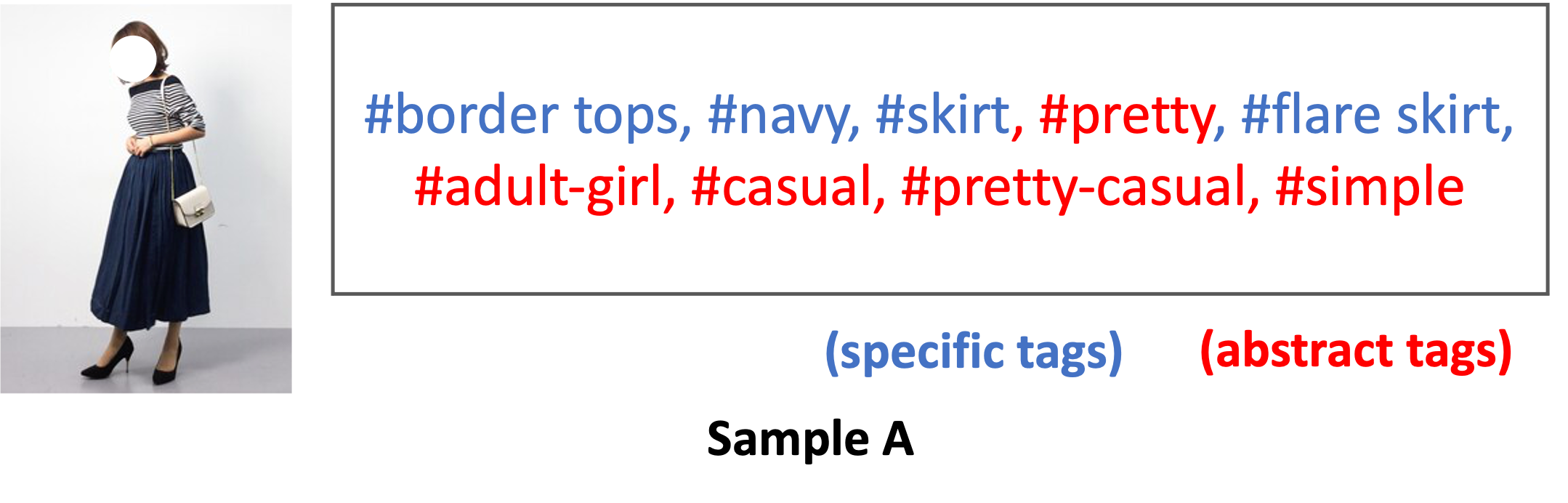}
\caption{Examples of samples in the target dataset~\cite{app_wear} \label{fig_sample_dataset}}
\end{figure}

The image data used in this study are full-body clothing photos of a single subject (a person). Each image is assigned several tags as attribute information from the user who posted the image. In addition, the tag information includes not only concrete and simple tags, such as ``border tops,'' ``navy,'' ``skirt,'' and ``flare skirt,'' but also abstract tags, such as ``pretty,'' ``adult-girl,'' ``casual,'' ``pretty-casual,'' and ``simple.''

One of its characteristics is that a specific tag, once attached, is always the correct tag regardless of the sensitivity of the contributor. In contrast, the characteristic of abstract tags is its uncertainties depending on the sensibility of the contributor. For instance, as per the sensibility of contributor A, if image A is completely ``pretty,'' the ``pretty'' tag can be attached by contributor A. Conversely, if contributor B feels that image A is only partially pretty, the ``pretty'' tag may not be attached by this contributor. In addition, for contributor C, if the expression ``cute'' seems more appropriate than ``pretty,'' the ``cute'' tag would be attached rather than ``pretty.'' Thus, a target full-body clothing image includes not only specific tags but also abstract tags. The abstract expressions are one of the major reasons why users find the fashion domain difficult.

\section{Methodology}
\subsection{Image Retrieval}
Images can be retrieved using image and tag adding or subtracting operations because the proposed model maps tags and images into the same projective space. Basic image retrieval is accomplished by adding (positive) and subtracting (negative) tags to the query image and is expressed as Eq.~(\ref{retrieval}).
\begin{eqnarray}
{\rm \mathbf{x}}_{\rm o} &=& \underset{{\rm \mathbf{x}}}{\rm argmax}\ s \left(
{\rm \mathbf{x}}_{\rm q} + {\rm \mathbf{v}}_{\rm pos} - {\rm \mathbf{v}}_{\rm neg}, {\rm \mathbf{x}}
\right), \label{retrieval}
\end{eqnarray}
where ${\rm \mathbf{x}}_{\rm o}, {\rm \mathbf{x}}_{\rm q} \in \mathbb{R}^{KL}$ denote the embedded representation of the output and query images, ${\rm \mathbf{v}}_{\rm pos}, {\rm \mathbf{v}}_{\rm neg} \in \mathbb{R}^{KL}$ are the embedded representation of the positive and negative tags, and $s(\mathbf{x}, \mathbf{y})$ indicates the cosine similarity between vectors $\mathbf{x}$ and $\mathbf{y}$.
This operation enables, for example, an image search for ``I want to know the coordination of office casual by subtracting the casual element from the target coordination.''

However, image retrieval based on the above calculation is a function that is also provided in the conventional Visual-Semantic Embedding (VSE) model~\cite{Shimizu2022_FashionIntelligenceSystem} and it cannot meet the detailed needs of users who want to make minor changes only to the upper clothes. In contrast, the proposed Partial VSE (PVSE) model allows the user to know to the parts each dimension in the embedded representation of images and tags corresponds to. Utilizing this advantage, delicate image retrieval is obtained such that changes are made only to the parts specified by the user and are given by adding or subtracting the embedded representation of the tag, as expressed in Eq.~(\ref{retrieval_part}).
\begin{eqnarray}
\hspace{-0mm} v_{{\rm pos}, k} &=& \left\{
\begin{array}{ll}
v_{{\rm pos}, k} & ({\rm if.} \; k \in {\rm K}_{\rm q}), \\
0.0	& ({\rm otherwise}),
\end{array}
\right. \label{retrieval_part}
\end{eqnarray}
where $v_{{\rm pos}, k}$ denotes the $k$-th dimension of ${\rm \mathbf{v}}_{\rm pos}$, and ${\rm K}_{\rm q}$ is the set of dimensions corresponding to the query parts (target parts to be modified) specified by the user. In addition, the negative tag ${\rm \mathbf{v}}_{{\rm neg}}$ is determined as expressed in Eq.~(\ref{retrieval_part}). Therefore, Eqs.~(\ref{retrieval}) and (\ref{retrieval_part}) realize image retrieval with focus on a specific part.

Furthermore, it was necessary to specify a positive tag and its corresponding negative tag to maintain the overall atmosphere of the query image in the conventional image and tag computation for retrieval using the VSE model.
However, it is possible to maintain the overall atmosphere by maintaining the embedding representation of dimensions corresponding to parts other than the specified parts with the embedding representation obtained from the proposed PVSE model.
Therefore, even without selecting a negative tag, Eqs.~(\ref{retrieval_part})--(\ref{retrieval2}) and (\ref{retrieval_part2}) enable image retrieval with minor changes made only to the specified part.
\begin{eqnarray}
{\rm \mathbf{x}}_{\rm o} &=& \underset{{\rm \mathbf{x}}}{\rm argmax}\ s \left(
{\rm \mathbf{x}}_{\rm q} + {\rm \mathbf{v}}_{\rm pos}, {\rm \mathbf{x}} \right), \label{retrieval2} \\
\hspace{-0mm} x_{{\rm q}, k} &=& \left\{
\begin{array}{ll}
0.0 & ({\rm if.} \; k \in {\rm K}_{\rm q}), \\
x_{{\rm q}, k} & ({\rm otherwise}),
\end{array}
\right. \label{retrieval_part2}
\end{eqnarray}
where $x_{{\rm q}, k}$ denotes the $k$-th dimension of ${\rm \mathbf{x}}_{\rm q}$.
Additionally, if there are multiple tags used to create ${\rm \mathbf{v}}_{\rm pos}$ (for instance, ``casual'' and ``khaki-colored'' upper clothes), the average of those tags is obtained and applied it to Eq.~(\ref{retrieval_part}) above.

\subsection{Image Reordering}
Because the proposed model maps words and images into the same projective space, the similarities (relevance scores) of all the images (to which the target tag is attached) to the target tag can be calculated, and the images are sorted in order of the scores. This function is also possible with the conventional VSE model. However, in this study, image reordering by focusing on a specific part can be obtained by calculating the relevance score of the images and target tags in only the features in the dimensions corresponding to the target part. This feature can be used, for instance, to respond to the natural desire of the user to ``look up a coordinated outfit with a particularly (not particularly) casual upper-clothes.''

\subsection{Attribute Activation Map Creation}
An attribute activation map (AAM) can be obtained using the VSE model by creating a heatmap of the relevance scores between the embedded representation corresponding to each grid and the specified tag.

Because each grid contains either a single or multiple parts, a weighting calculation using grid weight map $G^{'}_{(i,j)} = \{ g^{'}_{(i,j),1},\cdots,g^{'}_{(i,j),l},\cdots,g^{'}_{(i,j),L} \}$ is used to calculate the embedded representation corresponding to each grid, where $g^{'}_{(i,j),l} = N_{(i, j), l} / N_{(i, j)}$ and $N_{(i, j)}$ denote the number of pixels included in the $(i, j)$-th grid. Therefore, $g^{'}_{(i, j),l}$ is the fraction of pixels that contain the $l$-th part in all pixels in the $(i, j)$-th grid.

Eq.~(\ref{eq:grid_img_embedd}) expresses the embedded representation corresponding to the $(i,j)$-th grid ${\rm \mathbf{x}}_{(i,j)} \in \mathbb{R}^{K}$, considering the (single or) multiple parts included in the grid.
\begin{eqnarray}
\label{eq:grid_img_embedd}
{\rm \mathbf{x}}_{(i,j)} = \sum^{L}_{l=1} g^{'}_{(i,j),l} {\rm \mathbf{W}}_{{\rm I},{l}} {\rm \mathbf{f}}_{(i,j)},
\end{eqnarray}
The embedded representation of the tag to be compared with the $(i,j)$-th grid in image when calculating the relevance score ${\rm \mathbf{v}}_{(i,j)} \in \mathbb{R}^{K}$ is determined using Eq.~(\ref{eq:grid_tag_embedd}).
\begin{eqnarray}
\label{eq:grid_tag_embedd}
{\rm \mathbf{v}}_{(i,j)} &=& \sum^{L}_{l=1} g^{'}_{(i,j),l} {\rm \mathbf{v}}_{{\rm q}, l},
\end{eqnarray}
where ${\rm \mathbf{v}}_{{\rm q}, l} \in \mathbb{R}^{K}$ denotes the embedded representation of the $l$-th part of the query tag.

The relevance score in the $(i,j)$-th grid between the image and tag is obtained by calculating the similarity between ${\rm \mathbf{x}}_{(i,j)}$ and ${\rm \mathbf{v}}_{(i,j)}$.
It is possible to create an AAM while considering the ratio of each part reflected in each grid using this score.

\section{Experimental Evaluation}
\paragraph{Implementation Details}
The number of dimensions of the embedded representation for all comparison models is generally set at 128. In addition, in the loss function of Gaussian VSE (GVSE)~\cite{gvse}, the Euclidean distance is adopted, and the covariance matrix is a spherical matrix. In the loss function of dual Gaussian VSE (DGVSE)~\cite{Shimizu2022_DGVSE}, Kullback--Leibler (KL) divergence is adopted, and the covariance matrix is a spherical matrix. The number of encoder and decoder layers of transformer-VSE (TVSE)~\cite{transformer_vse} was set at three.

\section{Additional Analysis}
\begin{figure*}[ht]
\centering
\includegraphics[width=1.0\linewidth]{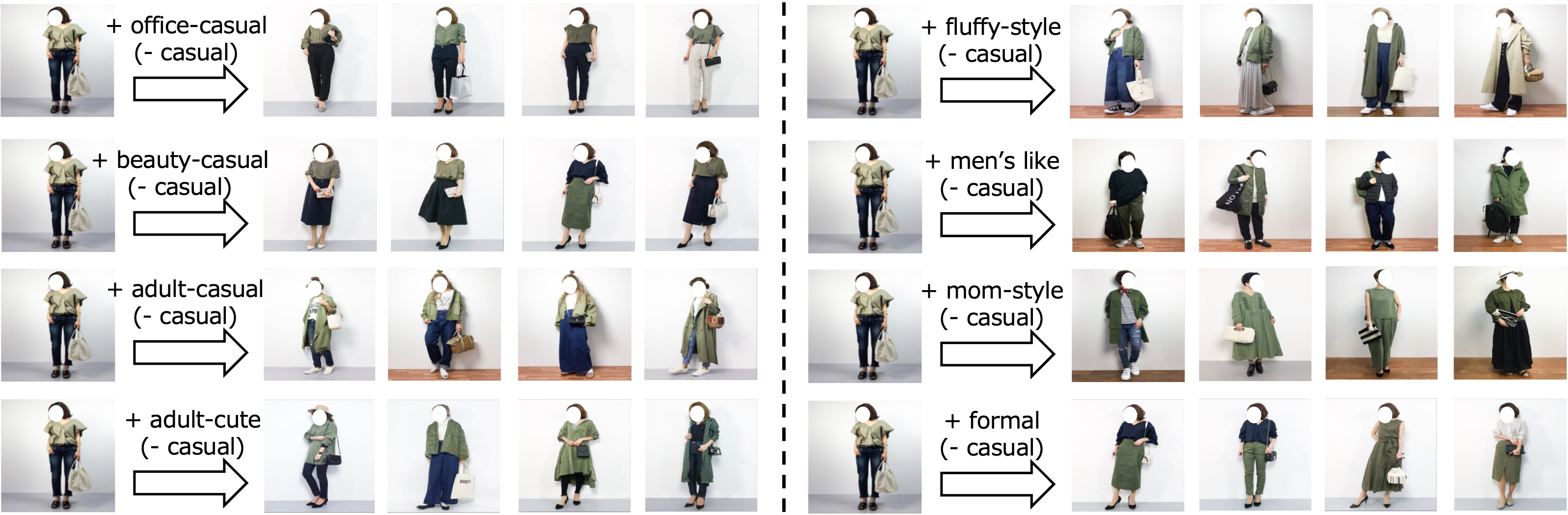}
\caption{Example of image retrieval by addition and subtraction for ``khaki'' and ``casual'' outfits \label{fig_retult_retrieval}}
\end{figure*}

\begin{figure*}[ht]
\centering
\includegraphics[width=1.0\linewidth]{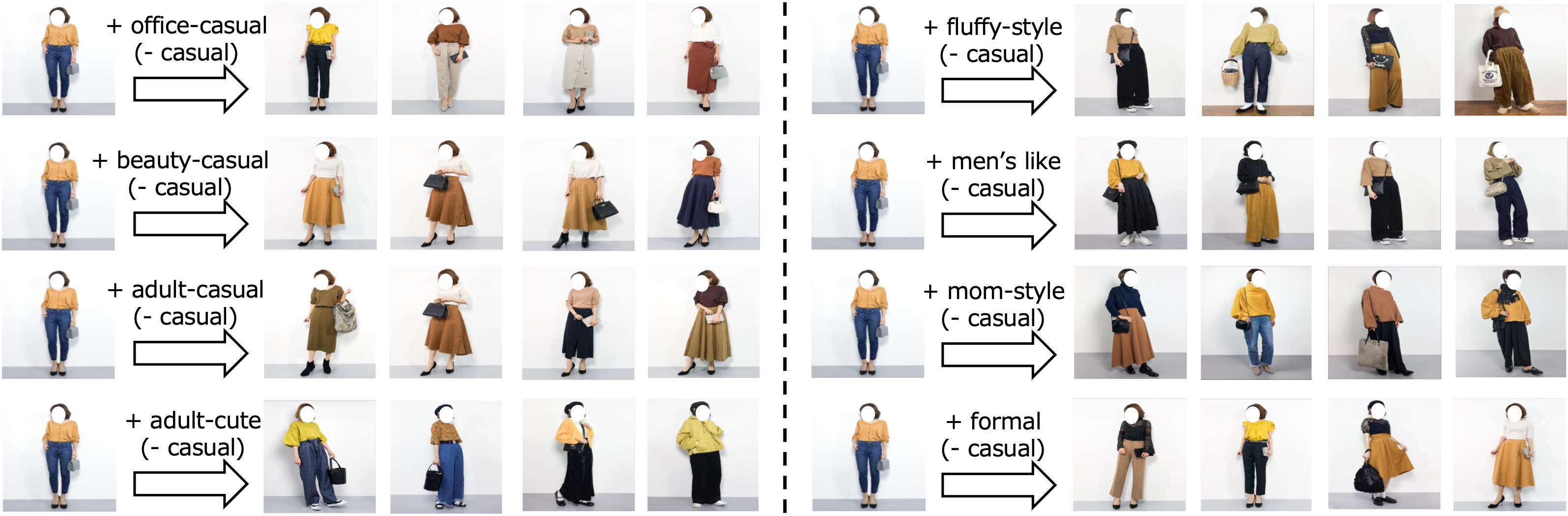}
\caption{Example of image retrieval by addition and subtraction for ``yellow'' and ``casual'' outfits \label{fig_retult_retrieval2}}
\end{figure*}

\paragraph{Image Retrieval and Reordering}
The results of the image retrieval obtained by image and tag addition and subtraction are shown in Figure~\ref{fig_retult_retrieval}-\ref{fig_retult_retrieval2}.

Thus, by using the search function, it is possible to easily grasp what kind of atmosphere is indicated by each ambiguous expression. This function allows users to search for various variations of clothing, such as ``casual,'' ``office-casual,'' ``beauty-casual,'' and ``adult-cute,'' while maintaining the user's preferred hue.

\section{Discussion}

\paragraph{Loss Function}
We were unable to include the experimental results for the evaluation of Experiment 2 because of space limitations; thus, they are summarized in Table~\ref{table:result_topk_aam_loss}.

\begin{table}[ht]
\centering
\caption{Summary of the evaluation values of the loss function type for top-$M$ regions selected using the similarity for each tag}
\label{table:result_topk_aam_loss}
\scalebox{0.65}{
\begin{tabular}{c|rr|rr|rr|rr}
\hline
\multicolumn{1}{l}{} &
  \multicolumn{2}{|c}{Head} &
  \multicolumn{2}{|c}{Upper-body} &
  \multicolumn{2}{|c}{Lower-body} &
  \multicolumn{2}{|c}{Shoes} \\
 &
  \multicolumn{1}{|c}{P@5} &
  \multicolumn{1}{c}{N@5} &
  \multicolumn{1}{|c}{P@5} &
  \multicolumn{1}{c}{N@5} &
  \multicolumn{1}{|c}{P@5} &
  \multicolumn{1}{c}{N@5} &
  \multicolumn{1}{|c}{P@5} &
  \multicolumn{1}{c}{N@5} \\ \hline \hline
triplet~\cite{triplet_loss}   & 0.161          & 0.145          & 0.547          & 0.494          & 0.346          & 0.306          & 0.096          & 0.087          \\
n-pair~\cite{npair_loss}      & 0.518          & 0.472          & 0.624          & 0.557          & 0.874          & 0.790          & 0.095          & 0.081          \\
single angular~\cite{angular_loss}   & 0.690          & 0.627          & 0.850          & 0.770          & 0.854          & 0.774          & 0.134          & 0.121          \\
batch angular~\cite{angular_loss} & 0.502          & 0.464          & 0.906          & 0.821          & 0.734          & 0.665          & 0.055          & 0.050          \\ \hline
n-pair angular~\cite{angular_loss} & \textbf{0.779} & \textbf{0.742} & \textbf{0.764}          & \textbf{0.691}          & \textbf{0.983}          & \textbf{0.888}          & \textbf{0.480}          & \textbf{0.463} \\ \hline
\end{tabular}
}
\end{table}

\end{document}